\documentclass[10pt,conference,a4paper]{IEEEtran}
\usepackage{booktabs} 
\usepackage{url}
\usepackage{soul,color,graphicx,float,epstopdf}
\usepackage{tablefootnote,textcomp,gensymb}
\usepackage{eurosym,booktabs,multirow}
\usepackage[caption=false]{subfig}
\usepackage{amsfonts} 
\usepackage{amsmath}
\usepackage{algorithmic}
\usepackage{array}
\usepackage{stfloats}
\usepackage{booktabs}
\usepackage{url}
\usepackage{float}
\usepackage{mathtools}
\usepackage{graphicx}
\usepackage{pdfpages}
\usepackage{subfig} 
\usepackage{balance}
\definecolor{Orange}{rgb}{1,0.5,0}

\begin{document}
\newpage
\title{Wireless Localisation in WiFi using Novel Deep Architectures}
\author{\IEEEauthorblockN{Peizheng Li\IEEEauthorrefmark{1}\IEEEauthorrefmark{2}\IEEEauthorrefmark{3},
Han Cui\IEEEauthorrefmark{3},
Aftab Khan\IEEEauthorrefmark{2},
Usman Raza\IEEEauthorrefmark{2},
Robert Piechocki\IEEEauthorrefmark{3},
Angela Doufexi\IEEEauthorrefmark{3},
Tim Farnham\IEEEauthorrefmark{2},
}\\ 
\IEEEauthorblockA{\IEEEauthorrefmark{2}Bristol Research \& Innovation Laboratory, Toshiba Europe Ltd., UK\\ 
\IEEEauthorrefmark{3}University of Bristol, UK\\
Email: {\{Peizheng.Li, Aftab.Khan, Usman.Raza, Tim.Farnham\}@toshiba-bril.com;}\\
{\{Han.Cui, R.J.Piechocki, A.Doufexi\}@bristol.ac.uk}}}

\maketitle
\begin{abstract}
This paper studies the indoor localisation of WiFi devices based on a commodity chipset and standard channel sounding. 
First, we present a novel shallow neural network (SNN) in which features are extracted from the channel state information (CSI) corresponding to WiFi subcarriers received on different antennas and used to train the model. The single-layer architecture of this localisation neural network makes it lightweight and easy-to-deploy on devices with stringent constraints on computational resources.
We further investigate for localisation the use of deep learning models and design novel architectures for convolutional neural network (CNN) and long-short term memory (LSTM).
We extensively evaluate these localisation algorithms for continuous tracking in indoor environments. 
Experimental results prove that even an SNN model, after a careful handcrafted feature extraction, can achieve accurate localisation. Meanwhile, using a well-organised architecture, the neural network models can be trained directly with raw data from the CSI and localisation features can be automatically extracted to achieve accurate position estimates. 
We also found that the performance of neural network-based methods are directly affected by the number of anchor access points (APs) regardless of their structure. 
With three APs, all neural network models proposed in this paper can obtain localisation accuracy of around 0.5 metres. 
In addition the proposed deep NN architecture reduces the data pre-processing time by 6.5 hours compared with a shallow NN using the data collected in our testbed. In the deployment phase, the inference time is also significantly reduced to 0.1 ms per sample. We also demonstrate the generalisation capability of the proposed method by evaluating models using different target movement characteristics to the ones in which they were trained. 
\end{abstract}

\begin{IEEEkeywords}
Wireless localisation, SNN, CNN, LSTM
\end{IEEEkeywords}

\section{Introduction}
The need for indoor location-based services in different sectors, such as smart homes, healthcare, smart industry, security, and agriculture demands a high level of accuracy and responsiveness. For instance, reliable and robust geo-fencing in security or safety-critical applications must detect unauthorised access and trigger quick response mechanisms in the event of zone breaches.  
Localisation can also provide real-time navigation, tracking and guidance services for people with impaired eyesight and the general public in indoor environments like museums, visitor attractions, shopping malls and other retail environments. 
However, achieving reliable indoor positioning is challenging due to multipath radio propagation which distorts the received signal properties, that are typically used in estimating the location, such as phase and amplitude.
Theoretically, the bandwidth of a radio signal determines the resolution and resilience to frequency selective fading of the corresponding localisation methods. Therefore,  Ultra-Wideband (UWB) technologies have an advantage in combating the multipath effects. 
However, UWB technologies are subject to different regional regulations on transmit power and spectrum use, which limits the usable bandwidths. Also, deployment of UWB infrastructure is costly as it is not normally used for other purposes. In contrast, the widely deployed commodity WiFi solutions use globally harmonised frequency bands and protocols, with higher bandwidths than Bluetooth and ZigBee technologies.  Thus, exploring signal properties at WiFi access points for positioning \cite{kotaru2015spotfi}  has gained popularity, making it one of the most attractive indoor localisation solutions. The problem of WiFi-based localisation is still in the effectiveness of information extraction in hostile multipath scenarios. Despite the massive amount of research effort in making commercially feasible indoor localisation schemes in the past decades, there is no clear winner that provides both high accuracy and low-cost deployment.

The orthogonal frequency-division multiplexing (OFDM) technique adopted in WiFi systems is useful in modelling signal transmission in indoor environments.
This scheme uses a large number of adjacent orthogonal subcarriers to divide the original wide-band channel into several narrowband channels. The channel state information (CSI) can be collected by receivers for every transmission at each carrier frequency \cite{ma2019wifi}.
Meanwhile, the release of CSI extraction tools \cite{halperin2010predictable,sen2013avoiding,xie2018precise} makes the processing and analysis of CSI possible. 
Currently, most notable WiFi localisation techniques are based on CSI because of its fine-grained representation of information compared to the received signal strength indicator (RSSI).
Different radiometric features such as the angle of arrival (AoA)  and time difference of arrival (TDOA) along with machine learning techniques may be utilised. These different features form the cornerstone of different localisation schemes.

In this paper, we study the indoor localisation of devices based on a commodity WiFi chipset. We utilise WiFi receivers within APs at fixed locations in a room and move a target transmitter device around while it broadcasts channel sounding packets to the receivers. Based on the CSI observed at the receivers, we present three novel localisation algorithms: a lightweight single-layer neural network using handcrafted features extracted from the CSI, a convolutional neural network (CNN), and a long short-term memory (LSTM) deep learning model using the raw CSI data directly.
In order to verify and evaluate the described algorithms, an optical tracking system \cite{optiTrack} was used to generate ground truth location coordinates to millimetre accuracy. The main research contributions of this paper are summarized below:
\begin{itemize}
\item To the best of our knowledge, this is the first work that applies a single layer shallow neural network (SNN) and demonstrates effective WiFi localisation. In order to do so, a novel method of extracting handcrafted localisation features from the raw radio CSI packets is employed.
\item This is the first application of deep learning (CNN and LSTM) models that directly utilise raw CSI packets as the training data. We demonstrate the packet combining methods and the design of corresponding neural network architectures.
\item Based on the data sets collected under the practical indoor scenario, the pros and cons of the above methods in terms of data processing complexity, temporal resolution and localisation accuracy are discussed thoroughly, as well as an ablation study is performed on the input data arrangement.
\item Deployment tests using pre-trained shallow and deep models are carried out and the generalisation performance is assessed using a data set collected with different target movement characteristics.
\end{itemize}


\section{Background}
\label{sec:Background}
\begin{figure}[t]
    \centering
    \includegraphics[width=0.38\textwidth]{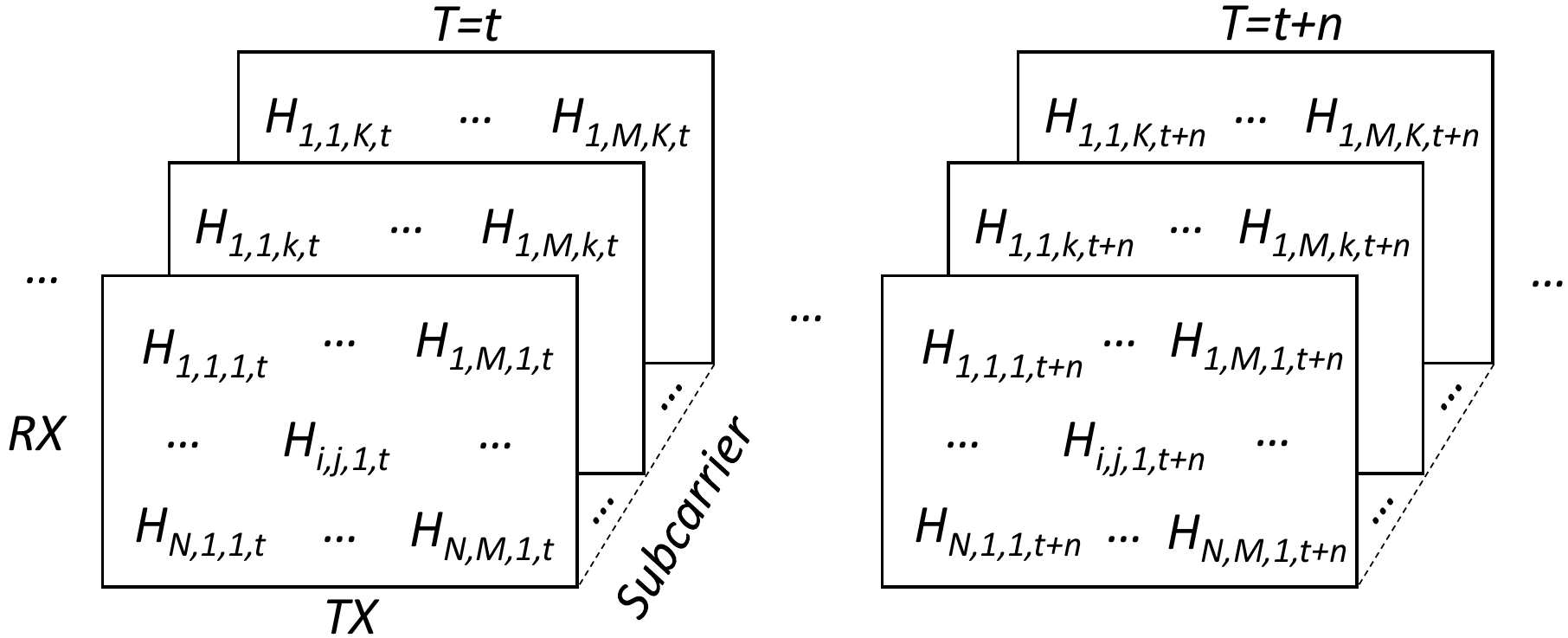}
    \caption{Indication of CSI tuples within a transmission time frame}
    \label{fig:CSI_matrix}
\end{figure}

Indoor localisation, tracking, and even human pose recognition based on WiFi signals have received much research attention recently. 
The theoretical basis for these localisation schemes is based on the inverse relationship between the received signal strength and the signal transmission distance.
However, due to the complex superposition of multiple signals travelling through different paths in the indoor environments, it becomes difficult to estimate distance accurately based on signal strength alone. 
Thus, early RSSI-based localisation schemes, such as RADAR \cite{bahl2000radar} and Horus \cite{youssef2005horus}, struggled to improve  accuracy to below  one metre due to the sensitivity of RSSI to the multipath effect in indoor environments. 
As Yang et al. \cite{yang2013rssi} showed, a stationary receiver experiences a 5 dB RSSI fluctuation in a typical laboratory environment over a short duration of one minute.  
We can distinguish between two classes of localisation enhancement scheme. 
The first class relies on the modification of hardware and the signal waveform. Examples include schemes that increase antenna diversity \cite{gjengset2014phaser} or use Frequency Modulated Continuous Wave (FMCW) \cite{adib2015capturing} using a single WiFi modulation frequency. 
The second class focuses on the development of algorithms based on information extracted from commodity chipsets using standard 802.11n channel sounding packets. Although special driver support is needed to extract the CSI data, but by doing so 
the use of fine-grained CSI becomes possible and opens  new research avenues for WiFi localisation. 
CSI characterises the propagation environment of wireless signals between transmitter and  receiver antenna pairs on the different subcarrier frequencies. Hence, the number of transmit and receive antennas (or sensors) and the number of subcarriers become important for capturing the multipath channel characteristics. 
CSI amplitudes and phases result from the combining of the direct and reflected signal paths through the multipath channel. Therefore, sufficient CSI subcarriers and sensors are needed to fully resolve the channel. In theory, each CSI entry represents the Channel Frequency Response (CFR):
\begin{equation}
H(f;t)=\sum\nolimits_{n}^{N}{{{a}_{i}}}(t){{e}^{-j2\pi f{{\tau }_{i}}(t)}}
\end{equation}
where $a_{i}(t)$ is the amplitude attenuation factor,  ${{\tau }_{i}}(t)$ is the propagation delay, and $f$ is the carrier frequency. 
From moment $T = t$ to $T=t+n$, an intuitive schematic diagram of CSI packets is shown in Figure \ref{fig:CSI_matrix}, where the number of transmitting antennas is $M$, the number of receiving antennas is $N$ and the number of subcarriers is $K$. 
An ideal CSI entry is a 3D tuple $H\in \mathbb{C}^{N\times M\times K}$.

The Intel 5300 NIC used in this paper can report $30$ subcarriers, so the CFR of subcarrier $i$ is:
\begin{equation}\label{formula: H}
{H}_{i}=|{H}_{i}|{{e}^{j\sin \{\angle {{H}_{i}}\}}, i\in \left [ 1,30 \right ] }
\end{equation}
where $|{H}_{i}|$ and $\angle {{H}_{i}}$ are the amplitude and the phase response of the subcarrier $i$ respectively. Correspondingly, amplitude and phase can be used as localisation parameters. For instance, FIFS \cite{xiao2012fifs} utilises a weighted average of the amplitude values over multiple antennas to achieve localisation. 
Phaser \cite{gjengset2014phaser} is the first attempt of the phase-based scheme. Meanwhile, in the case of trajectory tracking, these parameters can be used to calculate the Doppler velocity for developing localisation and tracking algorithms \cite{li2017indotrack}. 

The overall goal of the localisation scheme is to create a mapping between the expected CSI at each target location and its coordinates. 
Candidate solutions to this problem can be divided into three main categories; the first is based on the fingerprint method, which collects as many location-related CSI measurements as possible during the offline phase and extracts features to build a fingerprint database. During the on-line  deployment this data set is then used to perform the closest CSI lookup in order to match the received CSI feature to the most likely location, or set of $K$ nearest locations. 
However, due to its complex data collection and processing procedure, as well as workload associated with extensive manual surveys, it is difficult to use it in the real-world applications. 
The second approach is to extract received signal features, such as the AoAs estimated from the incident CSI, to achieve localisation through mathematical 
models \cite{kotaru2015spotfi} 
using geometric methods. An example is the triangulation of the AoA estimates from two or more known AP locations.  
The third category is based on machine learning, especially neural network (NN) models, due to their strong regression ability and flexible configuration. 
This particular approach has recently gained a lot of attention. 
However, there are specific challenges for such supervised approaches such as difficulties in collecting and labelling of complex training data sets, feature extraction, the structural design and training of the neural network.
In essence, NN-based localisation is an evolution of the fingerprint approach. 
Although both methods need to collect data offline and establish a database, researchers have shown that NN has a more ideal and general location prediction capability compared to the traditional fingerprint methods.

Early work of Deepfi \cite{wang2015deepfi} and Phasefi \cite{wang2015phasefi} utilised the amplitude and the phase information of subcarriers respectively, by training fully connected neural networks to realise localisation. In the recent work of Wang et al. \cite{wang2018deep}, the authors realised an accurate CNN localisation scheme by constructing the AoA estimates as the training set. Khatab et al. \cite{khatab2017fingerprint} proposed an auto-encoder based Deep Extreme Learning Machine indoor localisation method. Chen et al. \cite{chen2019wifi} proposes a localisation scheme using LSTM. 
However, these methods only discussed the structure and the training of a single network based on discrete data with a sparse location distribution. Optimised approach of training data generation was not considered that can establish a link to the received CSI property.
Moreover, the raw CSI packets have to be processed by certain algorithms such as AoA estimation \cite{wang2018deep}, or discrete wavelet transform (DWT) \cite{zhang2020indoor} in the pre-processing stage which makes it challenging in terms of a deployment that requires rapid processing.
\begin{figure}[t]
    \centering
    \includegraphics[width=0.39\textwidth]{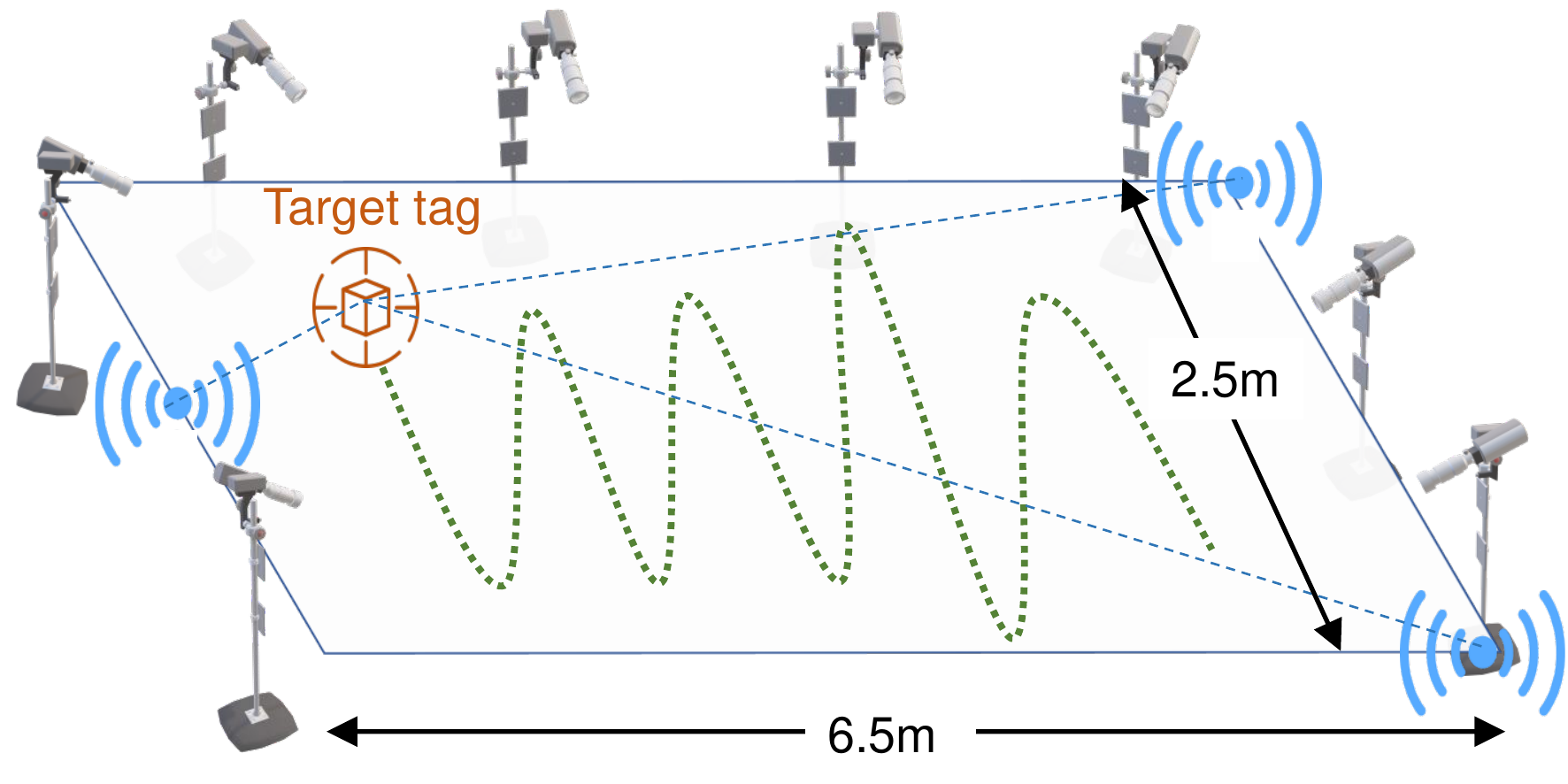}
    \caption{The experimental environment of WiFi data collection, where $8$ cameras constitute the OptiTrack system and provide the ground truth locations. The green dotted line represents the trajectory of the target tag.}
    \label{fig:experiment}
\end{figure}
\label{sec:exps}
In contrast to these developments, we use the raw CSI data directly as the input to train our deep neural network models, which to the best of our knowledge, is the first such attempt in the literature to do so. We present two novel deep neural networks models, the CNN model and the LSTM model, and describe the data processing procedure and the corresponding network architectures. We will show that neural networks trained with CSI data can achieve better localisation accuracy than a geometric, radio map assisted, method, described in \cite{farnham2019indoor}. Meanwhile, we also study the feasibility of combining the AoA estimation algorithm from \cite{kotaru2015spotfi} and a SNN. We present this novel combination as a reduced complexity approach compared to \cite{wang2017cifi}.

 \section{Experimental Setup and Dataset}
\label{sec:exp}
\subsection{OptiTrack Motion Capture Testbed}The experimental testbed for data collection is shown in Figure \ref{fig:experiment}. Three commodity WiFi APs are placed in the locations indicated in this figure. In the data collection period, these APs simultaneously receive broadcast channel sounding packets from the target transmitter, while the target tag makes curvilinear motions in the test area.  At the same time ground truth trajectory is recorded by an OptiTrack system in the form of Cartesian coordinates with a frame rate of $120$ Hz. 

\subsection{Dataset}  
The dataset is collected when the target object was carried by a walking human subject to mimic a severe shadowing environment and was continuously moved around in the test area. The dataset consists of five independent collection sessions, each lasting two minutes.
The channel sounding packet rate is set to $500$ Hz and CSI collected over $30$ subcarriers, with one transmit antenna and three receiver antennas at each AP. Because the sounding packet transmission rate ($500$ Hz) is higher than the frame rate of the OptiTrack system ($120$ Hz), interpolation and time synchronisation is necessary. In each data collection session, the channel sounding and the OptiTrack system tracking (the ground truth) start at the same time. After the collection, the OptiTrack data is interpolated to match the $500$ Hz rate of the CSI collected.
\section{Methods}
\label{NN models}
\subsection{Shallow Neural Network}
\label{subsec:SNN_method}
\begin{figure}[t]
    \centering
    \includegraphics[width=0.3\textwidth]{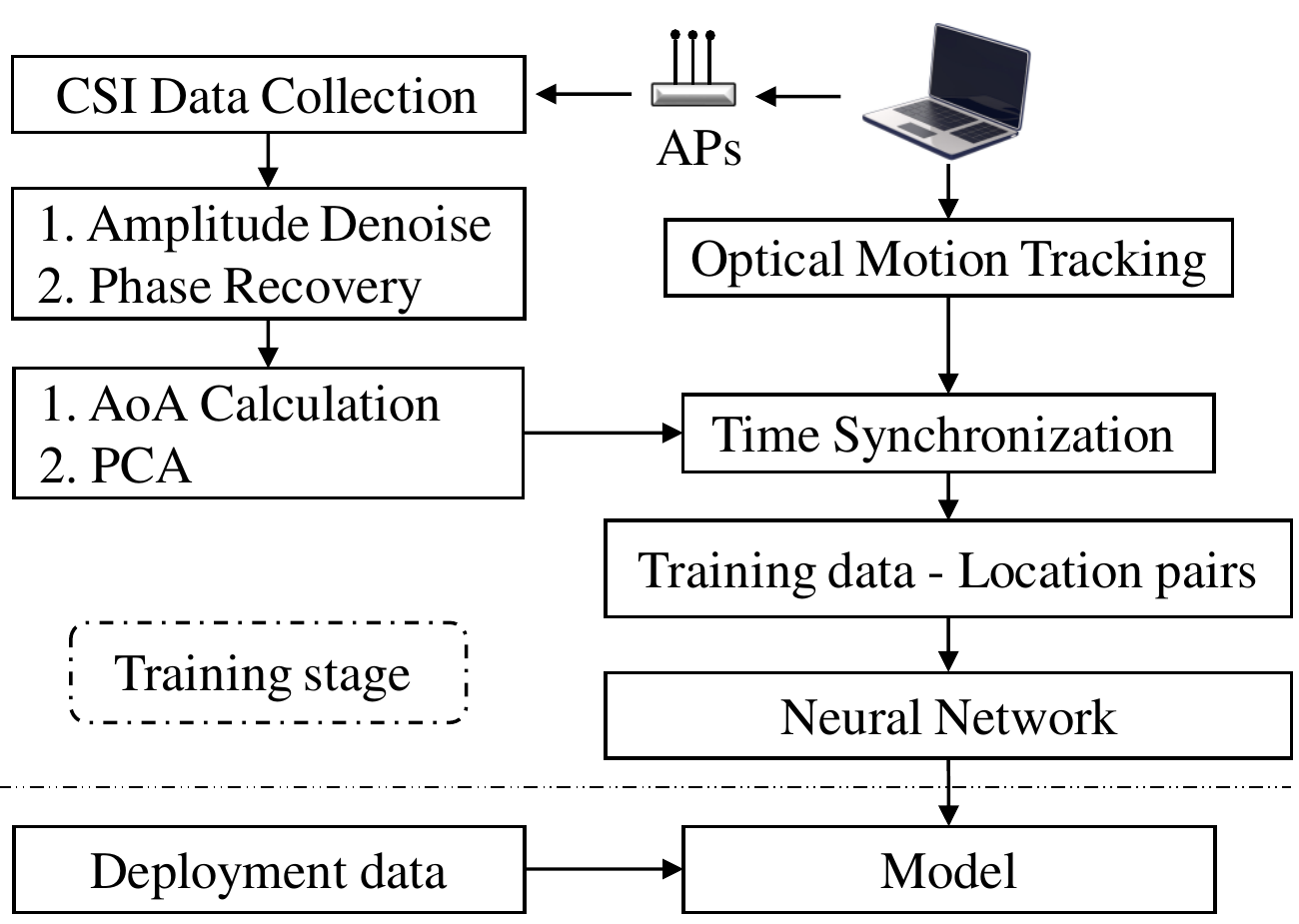}
    \caption{This figure shows the process diagram of WiFi data collection, feature extraction, training, and deployment of the SNN model.}
    \label{fig:wifi_shallow}
\end{figure}
 \begin{figure*}[t]
    \centering
    \includegraphics[width=0.80\textwidth]{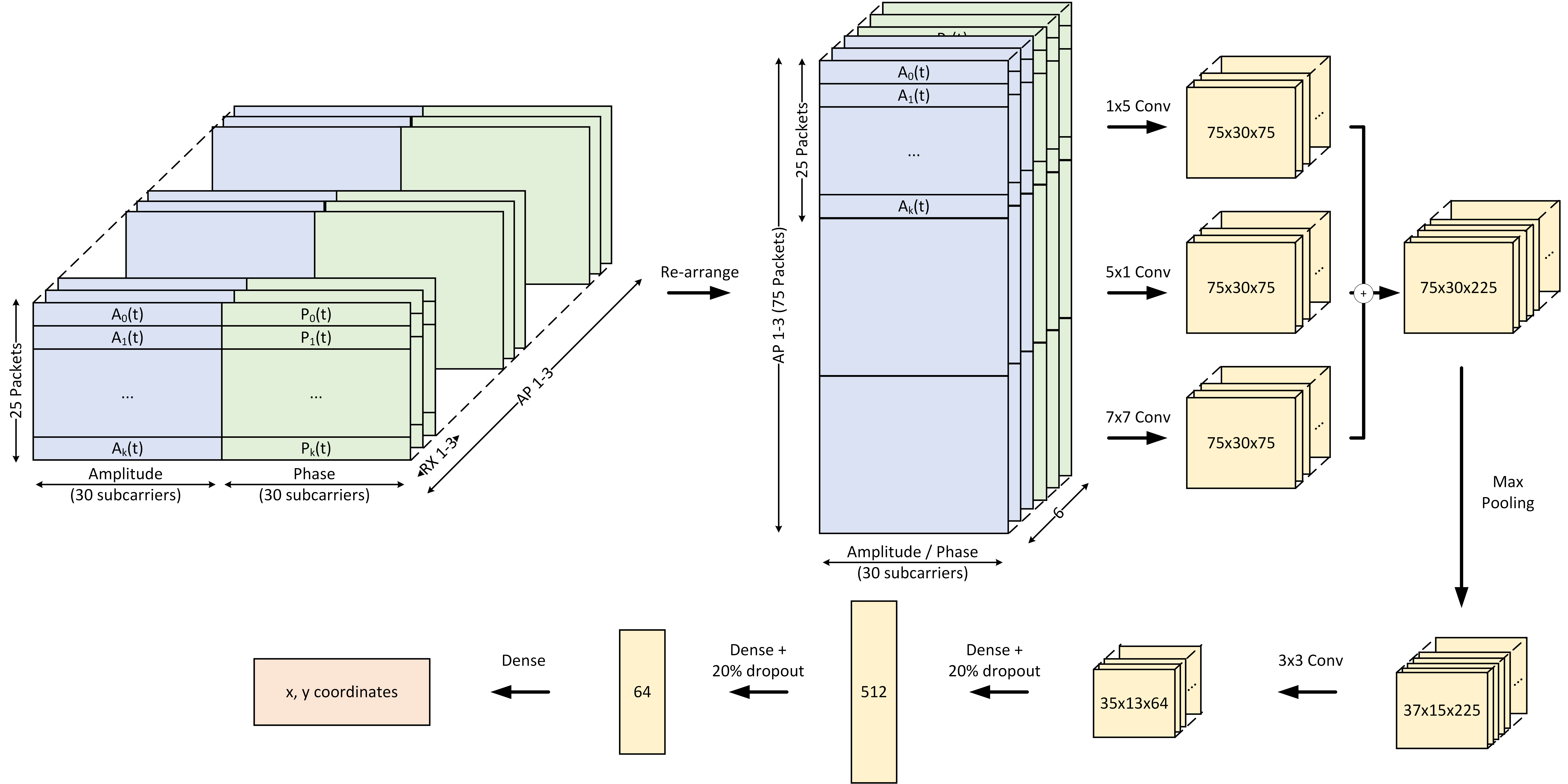}
    \caption{The CNN architecture and the training data structure. The array on the upper left represents the initial training data which is comprised of the raw CSI packets collected in all APs. The reshaped data is fed into the CNN model. The model outputs a corresponding $(x,y)$ coordinate of the target.}
    \label{fig:CNN_structure}
\end{figure*}
The proposed SNN localisation is inspired by the AoA based geometric approach, in which the location of the target can be deduced by the estimated incident signal angle at each anchor AP \cite{hou2018efficient}. In this process, the angle alignment calibration, filtering and smoothing are necessary to remove outliers.  This processing becomes more complex and computationally expensive when it is necessary to combine multiple angles and multiple APs. Therefore, we investigate the use of a SNN that processes the AoA data in order to perform localisation.

The small network size and the limited number of parameters in SNNs, result in a shorter training time and a computationally-efficient deployment for indoor localisation applications \cite{khan2019standing}. 
However, in the case of WiFi CSI, simple neural networks cannot perform effective signal extraction and accurately estimate target locations \cite{ma2019wifi}.
Therefore, CSI feature extraction must be performed and then fed into an SNN as input for training. Naturally, the AoA mentioned above can be a part of the training data. 
For each AP, we estimate the two most likely AoAs from each set of four consecutively received sounding packets.
Each estimated AoA is smoothed over successive time intervals, where the amount of smoothing is determined by the CSI transmission rate (one to two times higher than the transmission rate). However, the AoA alone is not enough to provide effective information, other parameters must be combined to achieve an accurate mapping.

For each AP, the CSI is divided into two parts corresponding to the amplitude and the phase of the subcarriers for each receiver antenna. For the amplitude of the received subcarriers any outliers, due to the hardware imperfection, need to be removed. In this paper we use a Hampel filter to remove the outliers.
The phase part of the CSI is not synchronised between the transmitter and receiver, only between antennas of the same AP, which needs to be corrected to restore the true phase.
The erroneous phase will have an unknown phase offset $\beta$ and time lag $\Delta t$.
 The phase of the channel response of subcarrier $f$ is $\angle \widehat{H}=\angle H+2\pi {{f}_{f}}\Delta t+\beta +{{Z}_{f}}$, where $\angle H$ is the genuine channel response phase and ${Z}_{f}$ represents the measurement noise.
We used a similar phase sanitisation method as mentioned in \cite{sen2012you}, that is,
\begin{equation}
\angle H=\angle \widehat{H}-\frac{\angle \widehat{{{H}_{F}}}-\angle \widehat{{{H}_{1}}}}{2\pi F}f-\frac{1}{F}\sum\limits_{1\le f\le F}{\angle \widehat{{{H}_{f}}}}
\label{formula:phase refine}
\end{equation}
After that, we perform the principal component analysis (PCA) \cite{wold1987principal} on the amplitude and phase of all subcarriers and keep the top five main components. As a result, for one AP, the five amplitude plus the five phase values after the PCA, and the two AoAs are constructed as the training data for SNN. 
For the three APs and the OptiTrack system adopted in this paper, as Figure \ref{fig:wifi_shallow} indicates, the  feature size of each training sample will be $36$ and the target variable will be the coordinates of the target at each timestamp.

For network training, we used a single layer architecture and the number of input nodes is set to $36$. We used $50$ hidden nodes and $2$ output nodes for a 2D Cartesian location estimation.
The two-dimensional output from the SNN is then post-processed with a moving average algorithm to smooth the result and produce the final estimation. 

\subsection{Convolutional Neural Network}
\label{subsec:CNN_method}
CNN has shown outstanding capabilities in object detection and classification and plays an increasingly important role in the area of image processing and beyond. 
A typical CNN architecture often consists of convolutional layers, pooling layers and fully connected layers. 
For CSI data, a single channel sounding packet corresponds to the natural image property and can be regarded as a multi-dimensional image. The problem of a single packet is that it carries limited information that is not easily extracted through convolution. 
Since the channel sounding transmission rate of the CSI data is relatively high ($500$ samples/s), a reasonable solution is to combine the data from multiple packets into one data item and reconstruct the dataset. 
In this paper, we chose to integrate $25$ packets into one training data subset and synchronise all the training data with the ground truth to generate complete feature-target pairs. 
Meanwhile, in order to exploit the temporal correlation between successive CSI packets via CNN, the re-arrangement of the combined data should preserve the inherent integrity of the CSI and make the features more obvious to the CNN. 
We set all the sub-carriers of each packet as the first dimension, the consecutively received $25$ packets in each AP as the second dimension, and the phases and amplitudes in the three antennas of each AP as the third dimension. Therefore, each input sample of the CNN becomes a $75\times30\times6$ tuple.
Before re-arranging this tuple, de-noising and sanitization are applied as discussed in Section \ref{subsec:SNN_method}. 
The data arrangement and the CNN structure are shown in Figure \ref{fig:CNN_structure}.
\begin{figure*}[t]
    \centering
    \includegraphics[width=0.80\textwidth]{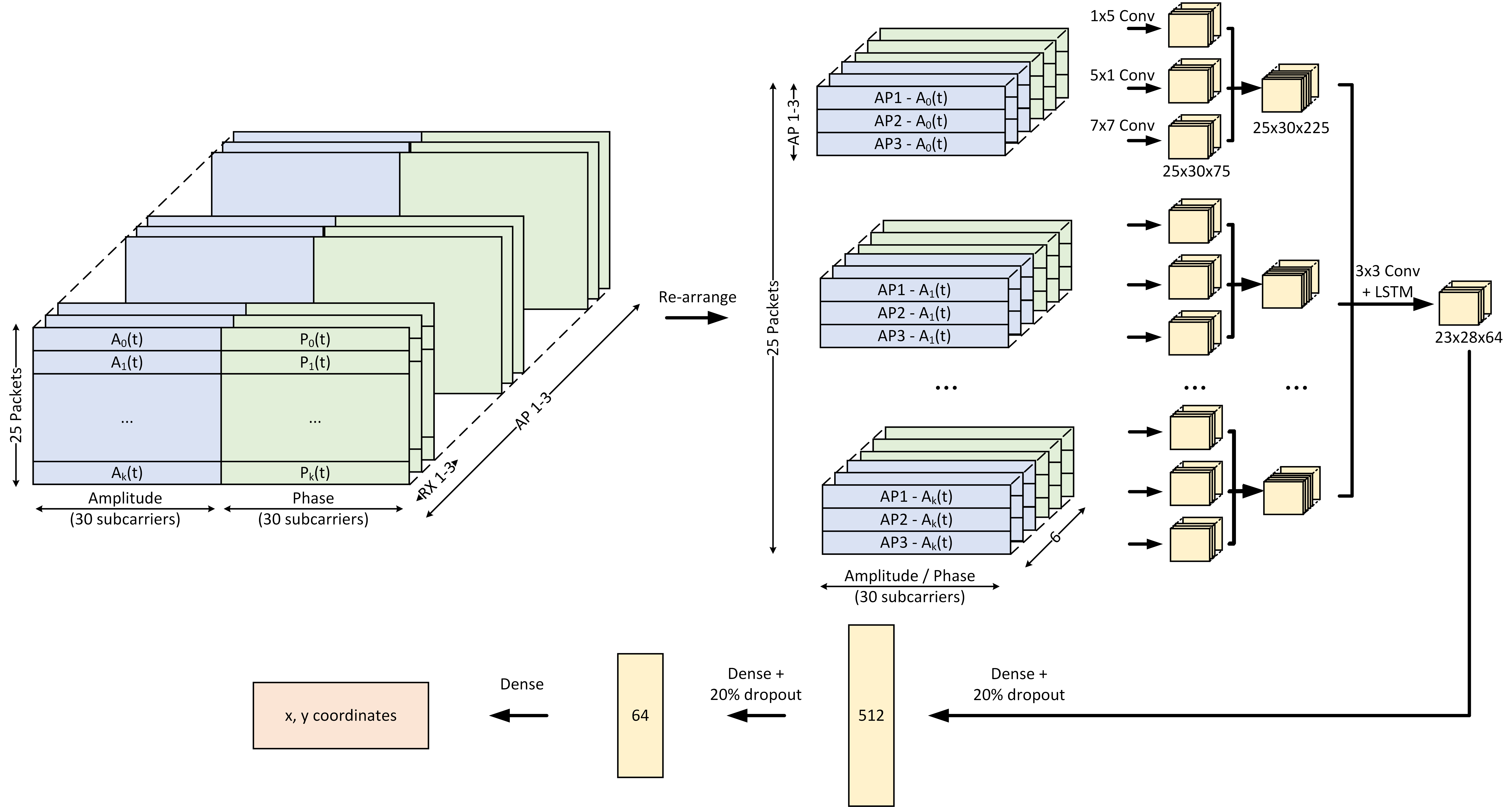}
    \caption{The LSTM architecture and the training data structure. The array on the upper left represents the initial training data which is comprised of the raw CSI collected in all APs. The reshaped data is fed into the convolutional layers and then the LSTM model. The convolution layers share the same structure as the CNN but the weights are re-trained. The two parts of the model are trained jointly and the output of this model is the $(x,y)$ coordinate of the target.}
    \label{fig:LSTM_structure}
\end{figure*}
The first step of the feature extraction is performed by applying a set of $1\times5$, $5\times1$, and $7\times7$ convolutional filters and the results are concatenated to include CSI information from different dimensions, where the horizontal and vertical filters are designed for adjacent CSI subcarriers and adjacent packets specifically. 
After the convolution, a max-pooling layer is applied to emphasise the features, which
produce a $37\times15\times225$ matrix, and then a $3\times3$ convolution layer is applied again. 
Three fully connected layers with size $512$, $64$ and $2$ are then applied. 
Between each two fully connected layers, a dropout layer is added with a $0.2$ drop rate to avoid over-fitting. 
The final output of this neural network is the $(x,y)$ coordinate of the target. We used the mean absolute error (MAE) as the loss function, the Adamax \cite{kingma2014adam} as the optimiser and the Scaled Exponential Linear Unit (SELU) as the activation function. The learning rate is set to $0.002$ and the batch size is $30$.



\subsection{Long Short-Term Memory}
\label{subsec:LSTM_method}
LSTM is a special structure of a recurrent neural network (RNN), which can solve the problem of long sequence dependence and diminishing or exploding gradient in RNN training \cite{pascanu2013difficulty}, and has a wide range of applications in sequence data processing like natural language processing (NLP). 
Since successive CSI packets collected can have hidden temporal correlation, the parameters in these packets can be processed as a long sequence language. Therefore, LSTM can also be used as the corresponding localisation method.

We use the same dataset as the CNN and the same initial data structure as previously proposed. 
That is, each training sample consists of the phases and amplitudes of 25 CSI packets, and the corresponding errors have been processed via the Hampel filter and the phase recovery method as discussed above. As shown in Figure \ref{fig:LSTM_structure}, we re-arrange the data sample into a time distributed subset, where every element in this subset consists of three CSI packets from the three APs respectively. 
Similarly, for each element, two convolutional layers are used to extract effective features, where the first layer is a combination of $1\times5$, $5\times1$ and $7\times7$ filters, and the second layer consists of $3\times3$ filters. 
Then three fully-connected layers with size $512$, $64$ and $2$ are used with a $0.2$ dropout in between.

\section{Results}
\label{sec:Discussion}
In this section, we provide evaluation results of the NN models using our datasets. We evaluate the models in terms of the overall localisation accuracy, the processing time, performance with different number of APs and the generalisation ability through a leave-one-session-out cross-validation scheme. Then we discuss the ablation study of input data arrangement.
Experiments are performed using the Intel i5-6500 CPU (each core running at $3.2$ GHz), $8$ GB of memory and an Nvidia RTX2070 Super GPU with $8$ GB memory. The data preprocessing is performed by the CPU whilst the training stage relies on the GPU. The training is based on Keras using TensorFlow 2.1 as the backend. 

\subsection{Performance Evaluation}
\label{subsec:Performance Evaluation}
We use the Euclidean distance between the estimated and the ground truth locations as an indicator for the localisation performance. 
That is, $e = \sqrt{(\hat{x}-x)^2+(\hat{y}-y)^2}$, where $(x,y)$ is the ground truth coordinate and $(\hat{x},\hat{y})$ is the final estimated coordinate. 
\begin{figure*}[t]   
    \subfloat[\label{fig:prediction_time}]{
      \begin{minipage}[t]{0.315\linewidth} 
        \includegraphics[width=2.1in]{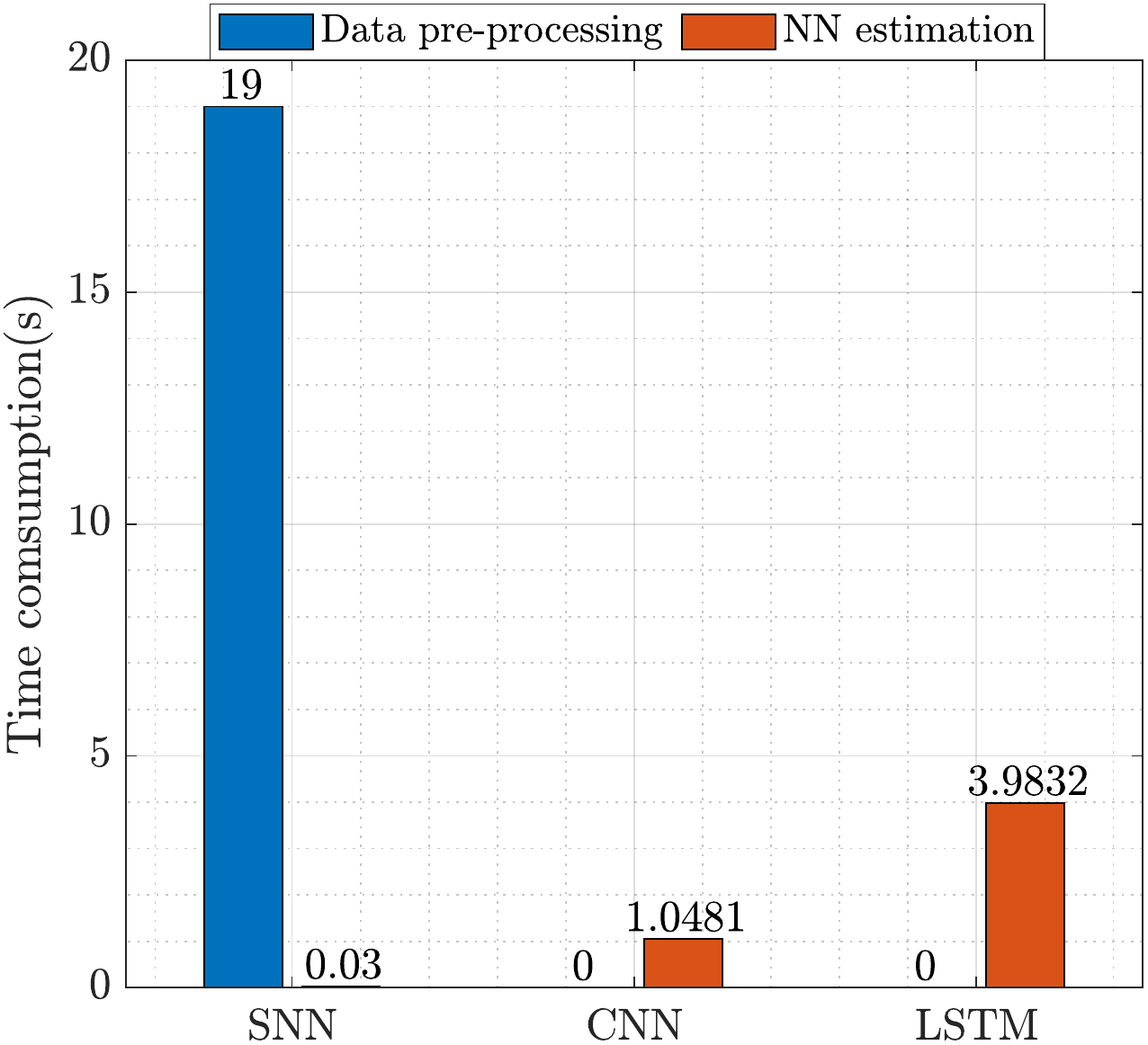}   
      \end{minipage}%
      }
      \hfill
        \subfloat[\label{fig:results of APs combination}]{
      \begin{minipage}[t]{0.315\linewidth}   
        \centering   
        \includegraphics[width=2.1in]{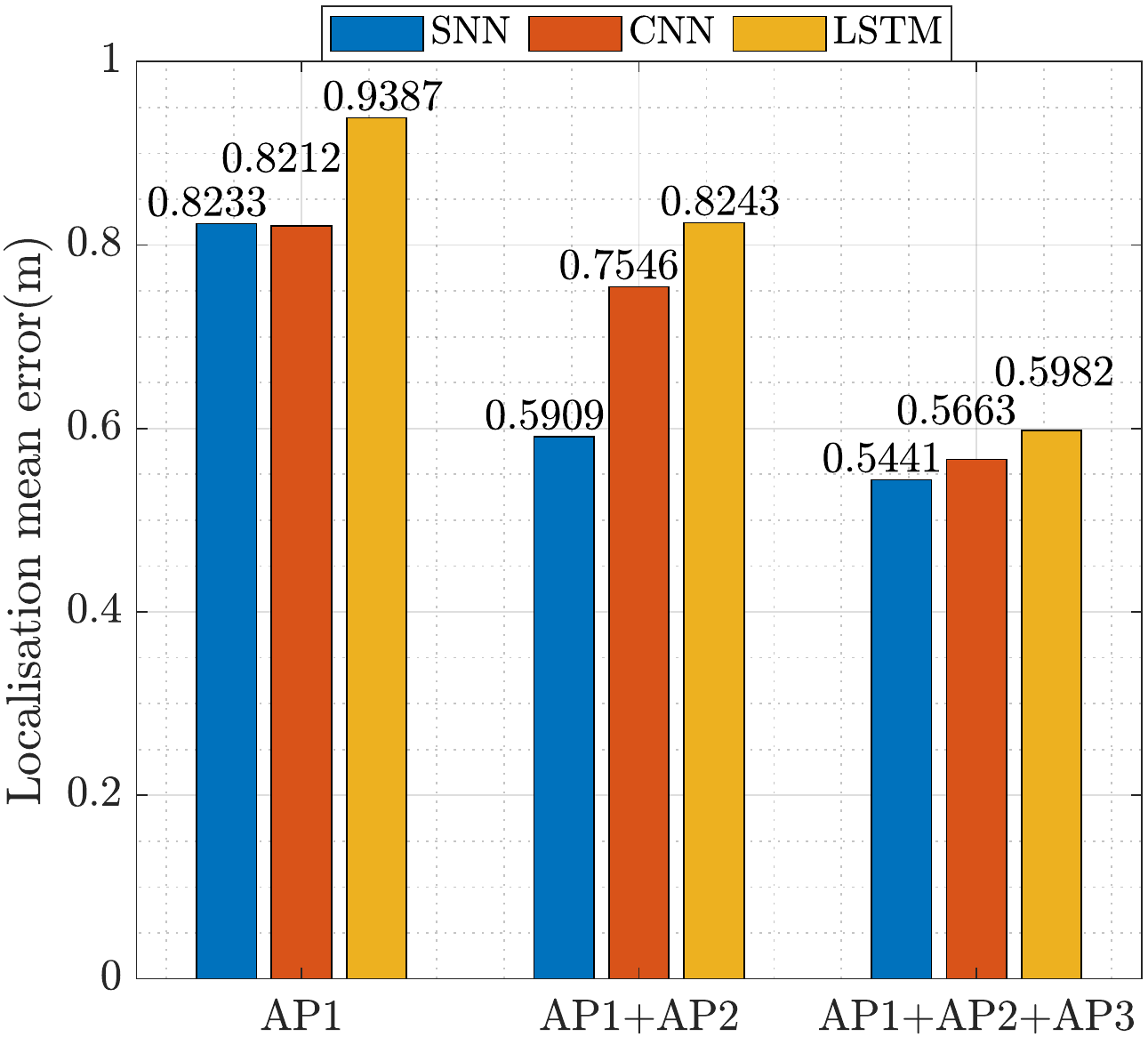}   
      \end{minipage} 
      }
     \hfill
     \subfloat[\label{fig:CNN structures compare}]{
      \begin{minipage}[t]{0.315\linewidth}   
        \centering   
        \includegraphics[width=2.1in]{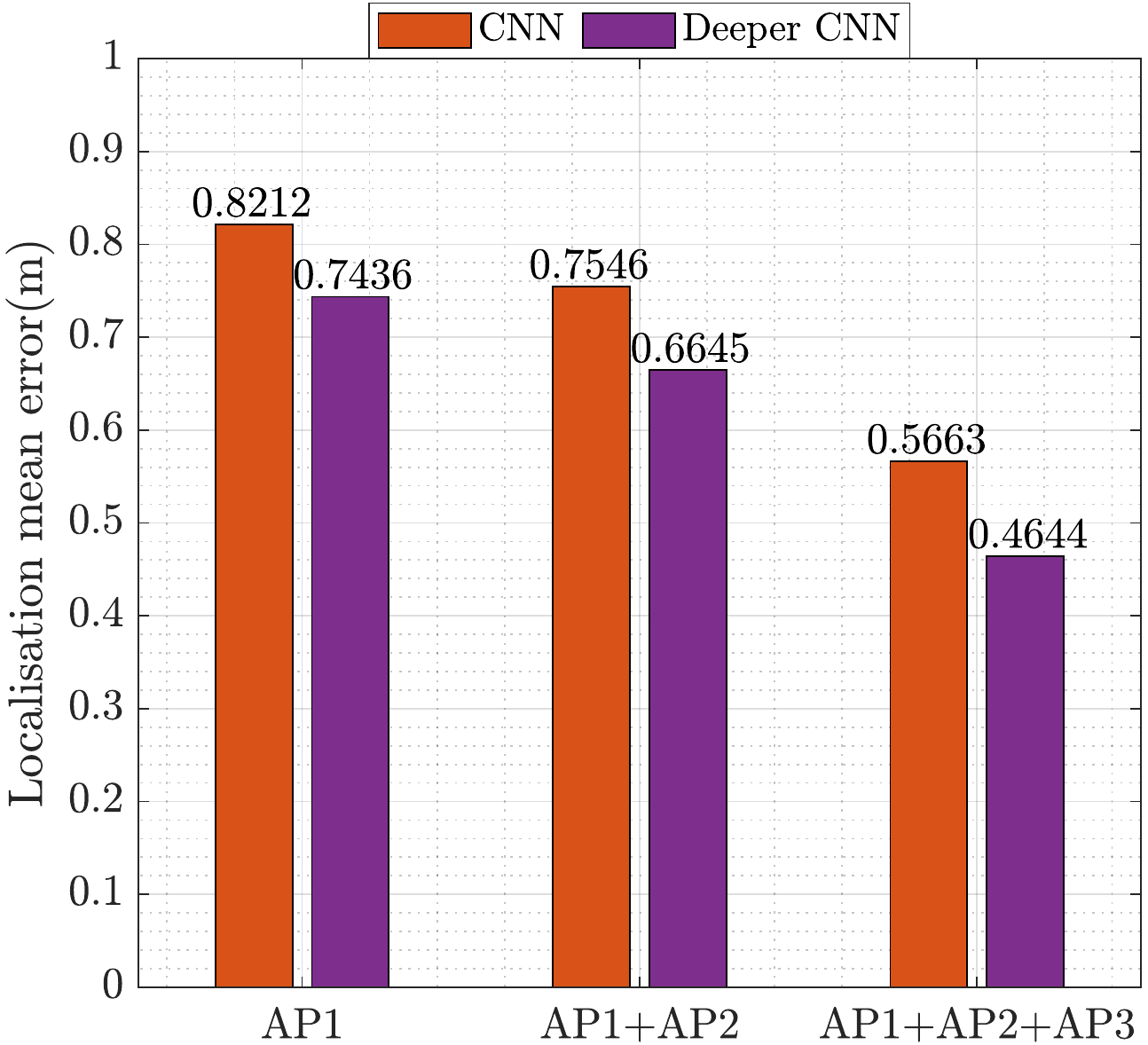}   
      \end{minipage}  
      }
      \caption{(a) Inference time comparison ($1000$ samples); (b) Accuracy comparison between the SNN, CNN and LSTM models under different AP combinations; (c) Accuracy comparison between the two CNN models under different AP combinations.} \label{fig:3tables}
    \end{figure*} 
{Table \ref{tab: mean error} gives the localisation error comparison of our proposed NN models.
In contrast to the SNN with handcrafted feature extraction, the deep CNN and LSTM can produce a comparable localisation accuracy with an automatic information extraction, where the mean error of the SNN, CNN and LSTM are $0.5441$~m, $0.5663$~m, and $0.5982$~m respectively.
When comparing to the work by Farnham \cite{farnham2019indoor} where a classical geometric localisation method is utilised, it can be seen that the performance of the proposed neural network approaches is better in non-line of sight situations. 
However, the referenced results in \cite{farnham2019indoor} relate to a two access point deployment and so, for fair comparison, the improvement is less significant but not negligible. Besides, unlike the geometric method \cite{farnham2019indoor}, NN approach does not need the extra AP's location or any additional independent calibration dataset.} 
\begin{table}[t]
\centering
\caption{Comparison of the localisation errors for different methods}
\label{tab: mean error}
\begin{tabular}{ccccc}
\toprule
        & Farnham \cite{farnham2019indoor}  & SNN & CNN    & LSTM   \\ \midrule
Mean(m)  & 0.75    & 0.5441     & 0.5663 & 0.5982 \\ \bottomrule
\end{tabular}
\end{table}
\subsection{Processing Time}
The comparison of the processing time consists of two parts. 
The first part is the data preparation and network training. The second part is the prediction, i.e., inference time after the system is deployed.
\subsubsection{Data Preparation and Training Time}
The main processing time for the SNN based localisation lies in the feature extraction, that is, the calculation of the AoA. 
Using the Music algorithm, an AoA is calculated every $4$ packets that takes about $0.11$ s on average. 
So for a single AP with a sampling rate of $500$ samples/s, it would take about $2.25$ hours to process $10$ minutes of data, and the processing of all the three APs would take $6.75$ hours.
Therefore, in the deployment stage of the SNN, the sampling rate of the channel sounding should be reduced to allow a sufficient AoA estimation processing time. Alternatively, the processing could be performed in parallel across multiple CPUs. 
The training time of this network is short with each epoch taking about $0.8$ s and after around $260$ epochs the network can converge to a desirable accuracy (around $0.55$ m), which takes $3.5$ minutes in total.
For the CNN and the LSTM model, the data pre-processing time is negligible, as only re-arrangement is required.
The average training time per epoch is around $6$ s for the CNN and $19$ s for the LSTM.
Both models were found to converge in around $100$ epochs, which took about $10$ and $32$ minutes respectively. 

\subsubsection{Inference Time}
The trained NNs can then be deployed for real-world applications. 
Taking the $1000$ collected samples as an example, the inference time of each network
 is shown in Figure \ref{fig:prediction_time}.
 The result indicates that, although the SNN is significantly faster than the CNN and the LSTM model, all three networks can be used in real-time processing, with the slowest LSTM model still being able to process data at higher than $250$ samples/s. 
 On the other hand, the data pre-processing time of the SNN takes most time and makes it less suitable for real-time localisation.


\subsection{Number of APs}
\label{Number of APs}
This section discusses the effect of the number of APs on the accuracy of the different neural network localisation algorithms. Recall that the location of each AP (AP1 to AP3) is shown in Figure \ref{fig:experiment}.
We kept the data structure and the architecture of each neural network unchanged, but changed the number of APs used and the amount of input data, in order to evaluate the importance of anchor APs. 
In the three cases of AP1, AP1 + AP2 and AP1 + AP2 + AP3, the accuracy of the SNN, CNN and LSTM models are shown in the Figure \ref{fig:results of APs combination}. 
It can be seen that, while all the algorithms are capable of working with different number of APs, having additional anchor APs is beneficial to improve the localisation accuracy. 
A single AP only has a localisation accuracy of about $0.8$ to $1$ m, while the combination of two APs has the accuracy increased to about $0.6$ to $0.8$ m, and the localisation of all three APs together have an accuracy of about $0.5$ m.
\begin{figure}[t]
    \centering
    \includegraphics[width=0.33\textwidth]{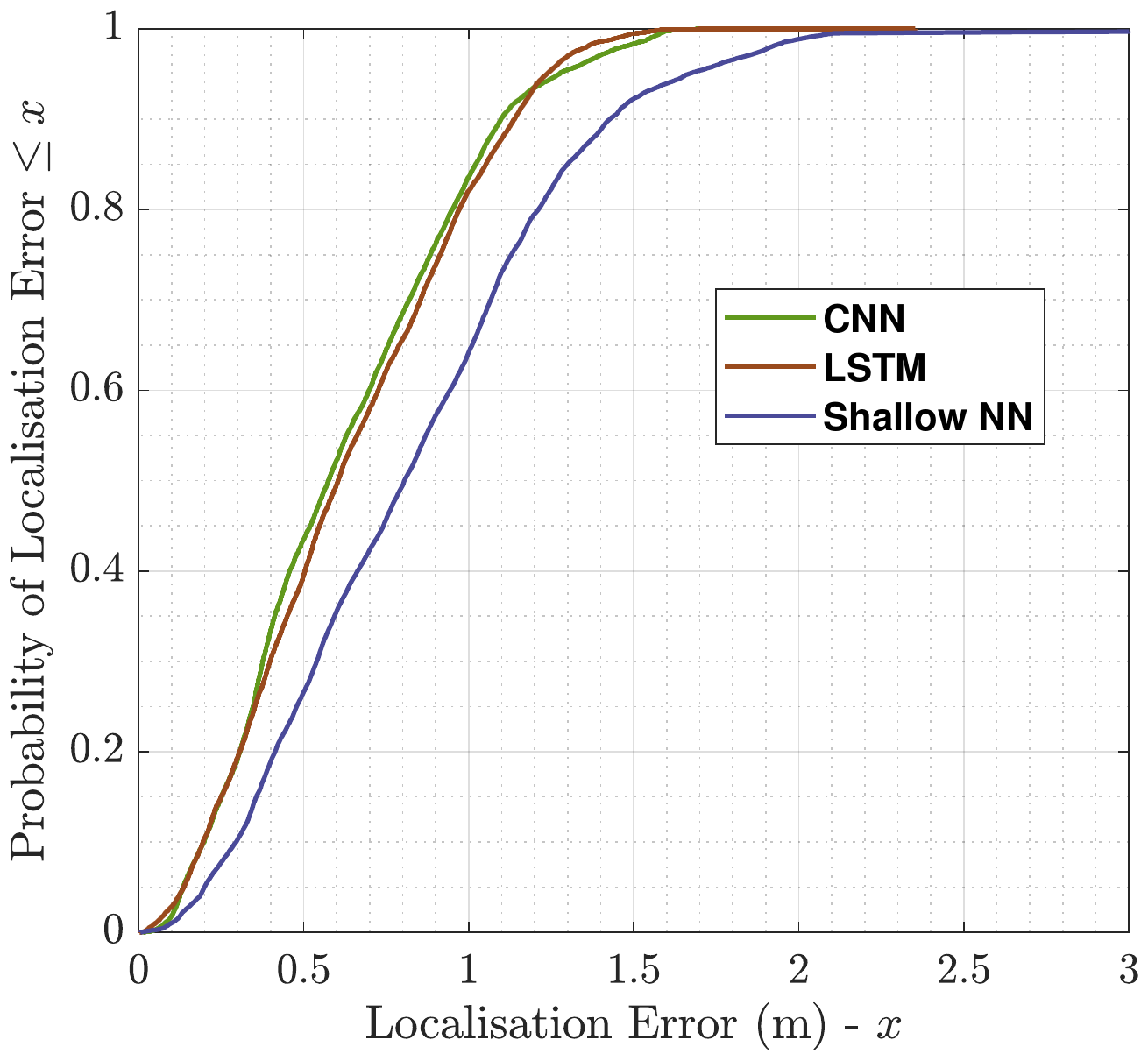}
    \caption{The localisation error CDF of the NN models in the special scenario. The LSTM model is similar to the CNN in accuracy and better than the SNN.}
    \label{fig:special_error}
\end{figure}
\begin{table}[t]
\centering
\caption{Localisation mean errors in the special movement scenario.}
\label{table:generalisation}
\begin{tabular}{cccc}
\toprule
               & SNN & CNN    & LSTM   \\ \midrule
Mean error (m) & 0.8559     & 0.6266 & 0.6421 \\ 
\bottomrule
\end{tabular}
\end{table}
\subsection{Special Use-case: Non-Constant Velocity Scenario}
All the experimental data is collected under the assumption of continuous motion at a constant speed as it is common in real-world applications. 
However, in order to evaluate the generalisation ability of the NN models we collected an additional set of data. In this, the operator carried the target, as before, but varied the speed of movement.
The motion characteristics contained a pattern sequence of linear-fast, stationary and linear-slow for two minutes. 
The CSI data collected is then used as the input to each of the trained neural networks described above for estimating the corresponding locations. The cumulative distribution functions (CDF) of the localisation error of the three network models are shown in Figure \ref{fig:special_error}, and the corresponding errors are given in Table \ref{table:generalisation}.

We note a relatively high error for the SNN. Moreover, the two-sample t-tests of SNN/CNN and SNN/LSTM indicate a significant difference. This is because when the velocity changes, the AoA estimation will have larger errors, which affects the input features in the SNN and therefore causes performance deterioration. 
For the CNN and LSTM, their relatively small errors indicate they have higher stability and are not significantly affected by the target's velocity.
We expect that the CNN and LSTM can have even better generalisation capabilities that can be further explored in the future. 

\subsection{Deeper CNN}
We have shown that our CNN model is capable of high accuracy localisation.
In this section, we study the possibility of further improving the accuracy with a deeper CNN structure. 
Based on the architecture in Figure \ref{fig:CNN_structure}, before and after max-polling, we added and concatenated a few more convolution filters of size $1\times1$, $1\times3$, $3\times1$ and $3\times3$ and increased the middle, fully connected layers with $100$, $80$, $60$ and $40$ nodes. 
This modification aims to improve the feature extraction ability of the CNN, at the expense of $2.8$ times more trainable parameters.
This deeper CNN is trained with the same dataset and method as in Section \ref{subsec:CNN_method}.
As a result, the training time of each epoch increased from $6$ s to $9$ s.
The final localisation mean error comparison is shown in Figure \ref{fig:CNN structures compare}. 
It is can be seen that the deeper CNN has better localisation accuracy performance in every scenario, especially for three anchor APs, where the localisation accuracy has improved by around $20\%$ (from $0.5663$ m to $0.4664$ m).
The trade-off between the complexity of the model and the performance allows the user to select the network design that best fits their application and resource.

We also attempted to modify the structure of the LSTM models using the above method.
However, in our experiments, the modified LSTM model did not show any significant localisation accuracy improvement.
\subsection{Data Arrangement Ablation Study}
\begin{figure}[t]
    \centering
    \includegraphics[width=0.365\textwidth]{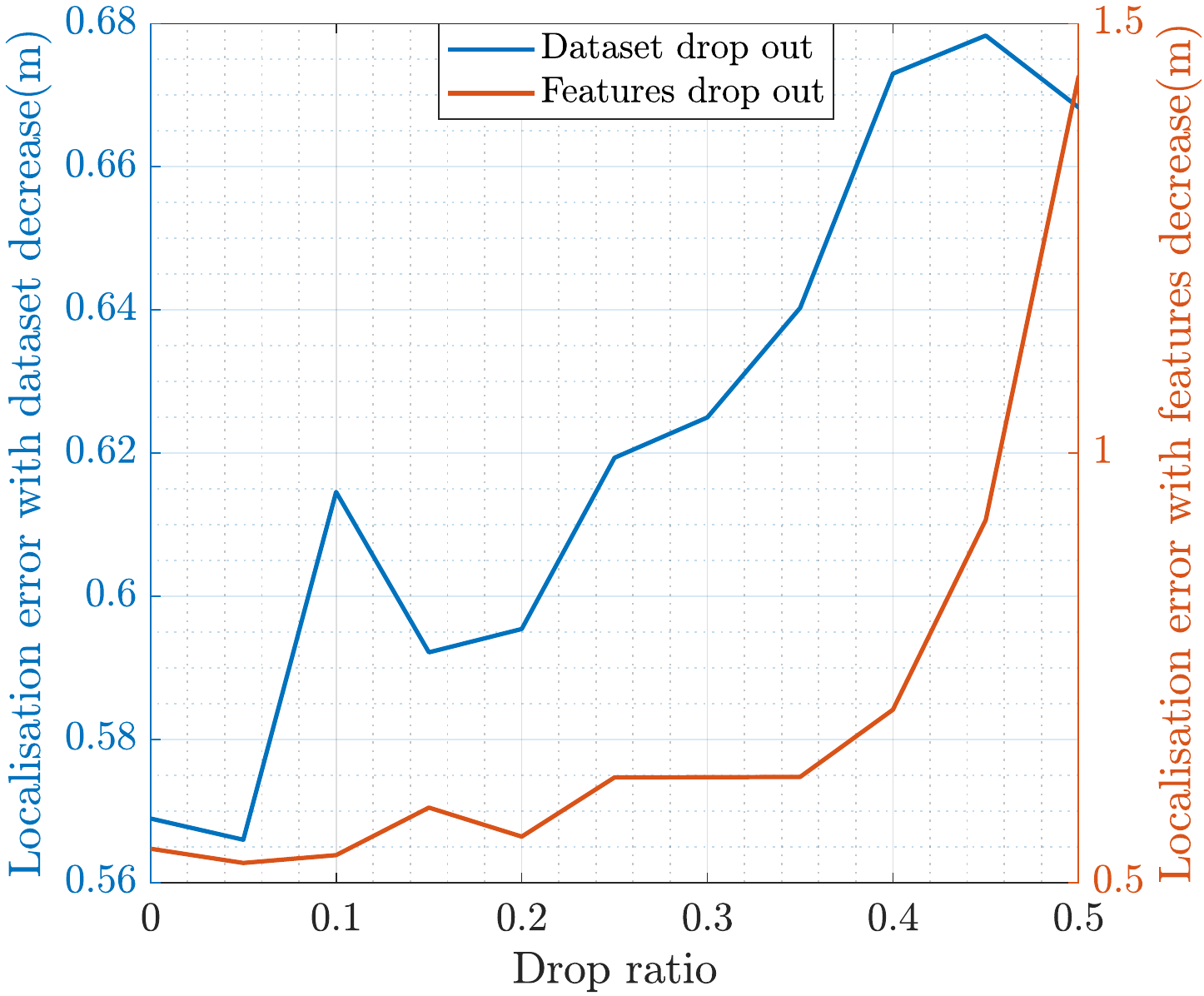}
    \caption{Ablation study for various training data (blue) and feature (red) sizes.
    }
    \label{fig:ablation}
\end{figure}
\begin{table}[t]
\centering
\caption{Features cropping ratio and network parameters relationship}
\label{tab: network size}
\resizebox{\columnwidth}{!}{%
\begin{tabular}{ccccccc}
\toprule
Cropping Ratio    & 0    & 0.1     & 0.2     & 0.3     & 0.4     & 0.5     \\ \midrule
Network Params & 100\% & 97.38\% & 94.77\% & 89.54\% & 86.92\% & 84.31\% \\ \bottomrule
\end{tabular}%
}
\end{table}
To assess the stability of the model with regards to the data arrangement, we performed two ablation tests on the training dataset and features. 
Firstly, after shuffling the training data set, we randomly selected a subset of the training data and dropped the others and trained the proposed CNN model using the selected data. For the second test, we selected the whole dataset, randomly deactivated a proportion of the training features to $0$, and trained the CNN model using the remaining features. The data drop rate was set between $0\%$ and $50\%$ during the experiments, whilst the test dataset was kept unchanged. 
Experiments were carried out $20$ times at each drop rate independently to minimise the impact of the random partitioning of the dataset. In Figure \ref{fig:ablation}, it can be seen that as the data size (blue line) and features (red line) are being dropped, the localisation accuracy decreases. 
The effect of deactivating the features is more obvious than dropping the data. Losing $50$\% of the training data or $40$\% of the features will reduce the localisation accuracy of the model by about $20$\%. 
We also explored the interplay of the features drop ratio and the number of associated trainable parameters of the model (as shown in Table \ref{tab: network size}). It indicates that the number of parameters is less impacted by the input feature size.
Therefore, we conclude that the majority of our training data and the features contribute to building a more accurate model and the current data arrangement is essential to ensure the stability of the model. 


\section{Conclusions}
\label{sec:Conclusion}
This paper describes several localisation algorithms based on commodity WiFi channel sounding and use of the resulting CSI data within Neural Network models. The focus has been on the principles of appropriate feature selection for SNN, CNN and LSTM, and their model architectures.
For the SNN, handcrafted localisation features are extracted from CSI using an AoA estimation algorithm and subcarrier PCA analysis, whereas CNN and LSTM are able to utilise the CSI raw data directly as the input. 
A novel indoor dataset was collected to evaluate the performance of the three algorithms. 
Our extensive evaluation shows that an increase in number of APs improves localisation accuracy of all the models. 
In the case of three anchors, the average localisation error is close to 0.5m for all algorithms. 
However, the manual information extraction stage of the SNN, i.e. the calculation of AoA, takes a significantly greater amount of time and therefore is less suitable for real-time  localisation applications.
On the other hand, the CNN and LSTM models show similar performance, however the LSTM model requires a longer training time. 
The advantage of the LSTM model is its ability to extract temporal information in the data which, we found, is less critical in this application. 
Finally, we show that the CNN and LSTM models have higher stability when the target is moving at a non-constant speed of motion. These novel insights can greatly help in designing future deep learning architectures for localization schemes for technologies other than WiFi. 
\bibliographystyle{IEEEtran} %
\balance
\bibliography{IEEEabrv,refs} 
\end{document}